\documentclass[twocolumn, switch]{article} 

\usepackage{preprint}

\usepackage{amsmath, amsthm, amssymb, amsfonts}

\usepackage[numbers,square]{natbib}
\usepackage{natbib}

\usepackage[utf8]{inputenc}	
\usepackage[T1]{fontenc}	
\usepackage{xcolor}		
\usepackage[colorlinks = true,
            linkcolor = purple,
            urlcolor  = blue,
            citecolor = cyan,
            anchorcolor = black]{hyperref}	
\usepackage{booktabs} 		
\usepackage{nicefrac}		
\usepackage{microtype}		
\usepackage{lineno}		
\usepackage{float}			
\usepackage{amsmath}
\usepackage{multirow}
\usepackage{lipsum}		

\usepackage{newfloat}
\DeclareFloatingEnvironment[name={Supplementary Figure}]{suppfigure}
\usepackage{sidecap}
\sidecaptionvpos{figure}{c}

\usepackage{titlesec}
\titlespacing\section{0pt}{12pt plus 3pt minus 3pt}{1pt plus 1pt minus 1pt}
\titlespacing\subsection{0pt}{10pt plus 3pt minus 3pt}{1pt plus 1pt minus 1pt}
\titlespacing\subsubsection{0pt}{8pt plus 3pt minus 3pt}{1pt plus 1pt minus 1pt}

\usepackage{tikz,xcolor,hyperref}
\usepackage{enumitem}

\definecolor{lime}{HTML}{A6CE39}
\DeclareRobustCommand{\orcidicon}{
	\begin{tikzpicture}
	\draw[lime, fill=lime] (0,0)
	circle [radius=0.16]
	node[white] {{\fontfamily{qag}\selectfont \tiny ID}};
	\draw[white, fill=white] (-0.0625,0.095)
	circle [radius=0.007];
	\end{tikzpicture}
	\hspace{-2mm}
}
\foreach \x in {A, ..., Z}{\expandafter\xdef\csname orcid\x\endcsname{\noexpand\href{https://orcid.org/\csname orcidauthor\x\endcsname}
			{\noexpand\orcidicon}}
}

\title{LokiTalk: Learning Fine-Grained and Generalizable Correspondences to Enhance NeRF-based Talking Head Synthesis}

\usepackage{authblk}

\author[1]{Tianqi Li}
\author[1\dag]{Ruobing Zheng}
\author[2]{Bonan Li}
\author[2]{Zicheng Zhang}
\author[1]{Meng Wang}
\author[1]{Jingdong Chen}
\author[1]{Ming Yang}

\affil[1]{Ant Group}
\affil[2]{University of Chinese Academy of Sciences}
\affil[ ]{\texttt {\{shijian.ltq,zhengruobing.zrb,jingdongchen.cjd,m.yang\}@antgroup.com}}

\begin{document}

\twocolumn[ 
  \begin{@twocolumnfalse} 

\maketitle

\centering{\textbf{Project Page:} \href{https://digital-avatar.github.io/ai/LokiTalk/}{\textcolor{red}{https://digital-avatar.github.io/ai/LokiTalk/}}}

\hspace{\fill}

\begin{abstract}
  Despite significant progress in talking head synthesis since the introduction of Neural Radiance Fields (NeRF), visual artifacts and high training costs persist as major obstacles to large-scale commercial adoption. We propose that identifying and establishing fine-grained and generalizable correspondences between driving signals and generated results can simultaneously resolve both problems. Here we present LokiTalk, a novel framework designed to enhance NeRF-based talking heads with lifelike facial dynamics and improved training efficiency. To achieve fine-grained correspondences, we introduce Region-Specific Deformation Fields, which decompose the overall portrait motion into lip movements, eye blinking, head pose, and torso movements. By hierarchically modeling the driving signals and their associated regions through two cascaded deformation fields, we significantly improve dynamic accuracy and minimize synthetic artifacts. Furthermore, we propose ID-Aware Knowledge Transfer, a plug-and-play module that learns generalizable dynamic and static correspondences from multi-identity videos, while simultaneously extracting ID-specific dynamic and static features to refine the depiction of individual characters. Comprehensive evaluations demonstrate that LokiTalk delivers superior high-fidelity results and training efficiency compared to previous methods. The code will be released upon acceptance.
\end{abstract}
\vspace{0.35cm}

  \end{@twocolumnfalse} 
] 


\let\thefootnote\relax\footnotetext{\dag ~Corresponding Author}


\section{Introduction}
The creation of realistic talking head videos from audio input has recently become a significant area of research, offering wide-ranging applications in fields such as video production, virtual assistants, and television commerce. These methods generally fall into two categories. The first category methods~\cite{kumar2017obamanet,tian2019audio2face,zheng2021neural} involves training on videos of a single individual, which can capture more individual characteristics. The second category focuses on one-shot talking heads ~\cite{hong2022depth,zhang2023sadtalker,cheng2022videoretalking,tian2024emo}. These methods require only a single photograph of the target individual, with facial movements entirely generated from audio inputs, thereby lacking personal distinctiveness. In this paper, we focus on the first category of methods based on individual characters.

Early  Generative Adversarial Network (GAN) based methods~\cite{ chen2020talking, suwajanakorn2017synthesizing} dominated the field of single-identity talking head generation. These approaches typically involved learning a temporal mapping between audio features and intermediate representations of facial movements~\cite{thies2020neural, zheng2021learning}, and generated photorealistic videos~\cite{prajwal2020lip, zhou2019talking}. However, GAN-based methods often struggled to maintain consistent identities across different frames, resulting in issues such as varying tooth sizes and fluctuating lip thickness~\cite{peng2024synctalk}. Furthermore, these methods tended to produce noticeable distortions and artifacts when handling significant changes in facial expressions or head poses. 

Recent advancements in Neural Radiance Fields (NeRF) techniques~\cite{mildenhall2021nerf} have significantly enhanced the quality of talking head generation. Benefiting from a unified implicit 3D representation, NeRF-based talking head~\cite{guo2021ad, tang2022real, li2023efficient} have demonstrated advantages in terms of multi-view 3D consistency, identity consistency, and facial details~\cite{yao2022dfa,liu2022semantic, shen2022learning}. Despite these advantages, NeRF-based methods also face critical challenges limiting their widespread commercial application. We elaborate on this from two perspectives:

\begin{itemize}[leftmargin=20pt]
\item Vsisual artifacts. The primary issues include inaccurate lip sync, unnatural expressions, disorderly blinking, unstable head motion, and disconnected head-torso movements. We attribute these issues to the lack of precise mapping between driving signals and their corresponding regions of influence. For instance, speech directly determines lip movements, while head motion and blinking have weaker correlations. The deformation of the torso is primarily driven by the head posture, while simultaneously being influenced by the jaw movements caused by lip motions. Consequently, the relationship between driving signals and portrait motions is complex and hierarchical. The simplistic learning of an audio-to-overall mapping can result in various visual artifacts.

\item Training efficiency. NeRF-based methods demand relatively more in terms of training time, memory usage, and data requirements~\cite{shen2022learning} compared to GAN-based approaches. This substantially increases costs for large-scale commercial applications. We posit that the primary reason for this is that the model completely relearns facial geometry and motion characteristics for each individual. However, talking portrait data contains considerable common information in both static and dynamic patterns, with personal characteristics merely being extensions of these common features.
\end{itemize}

In response to the aforementioned dual challenges, we propose LokiTalk, a novel approach that learns fine-grained and generalizable correspondences between driving signals and generated results, thereby enhancing NeRF-based talking head synthesis with lifelike facial dynamics while improving training efficiency.

\begin{itemize}[leftmargin=20pt]

\item To minimize visual artifacts, we identify and establish fine-grained correspondences. We introduce region-specific deformation fields, which explicitly decompose the overall talking portrait motion into distinct components, including facial and lip motion, eye blinking, head movements, and torso movements. By hierarchically modeling the driving signals and their associated regions through two cascaded deformation fields and optimizing within a unified canonical space, we significantly improve dynamic accuracy while minimizing synthetic artifacts.

\item To improve training efficiency, we learn generalizable correspondences shared among different individuals, followed by personalization based on individual characteristics. We introduce ID-Aware Knowledge Transfer, a pre-trained, plug-and-play module designed to augment NeRF-based methods. Through deliberate design, this module can learn universal static geometric representations and dynamic motion patterns from a limited number of high-quality multi-identity videos, while simultaneously extracting ID-aware static and dynamic features to refine the depiction of individual characters. 
\end{itemize}

Extensive experimental results demonstrate that LokiTalk produces high-fidelity talking portraits with notable efficiency and realism. We also show the generalizability of our method through a recent representative approach~\cite{li2023efficient}, which achieves comparable performance while reducing the number of training steps and data requirements.

\begin{figure*}[t]
  \centering
  \includegraphics[width=1\linewidth]{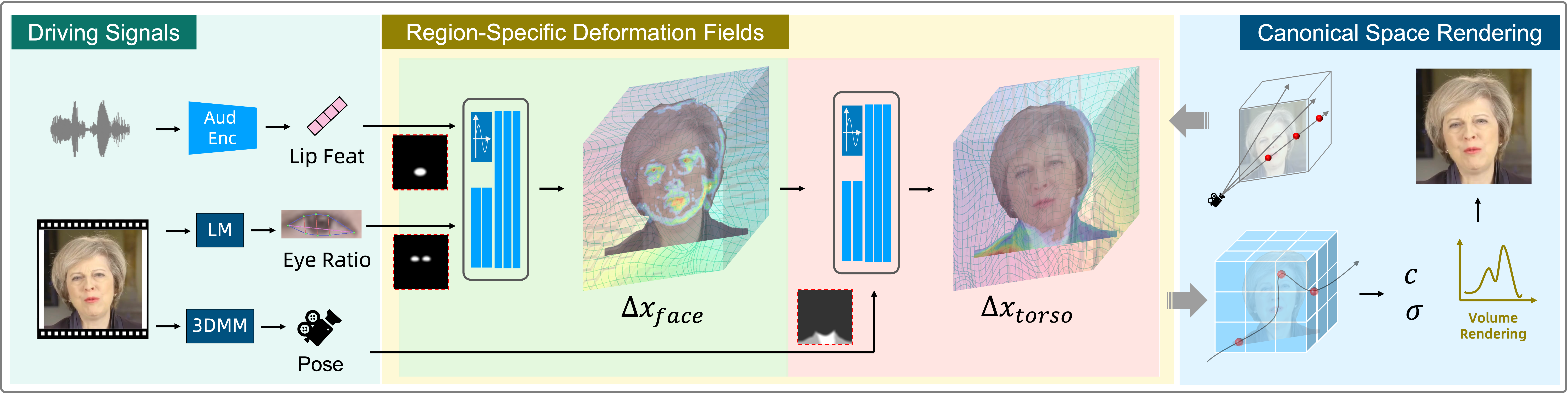}
  \caption{\textbf{Overview of the proposed Region-Specific Deformation Fields.} The driving signals (audio, pose, eye ratio) participate in the two-stage prediction of face and torso deformation fields, respectively. The mask subsequent to each driving signal represents the cross-attention loss between the driving signal and the corresponding region. A colored cubic grid is used to illustrate the predicted deformation fields, with the internal heat maps indicating the magnitude of the deformation amplitude.}
  \label{fig:framework}
\end{figure*}

\section{Related Work}

Neural Radiance Fields (NeRF)~\cite{mildenhall2021nerf} have emerged as a highly effective means of representing 3D objects. Unlike earlier methods that require 3D supervision, recent advancements in differentiable rendering techniques~\cite{zhang2024learning} have enabled direct training from images. Notably, NeRF has shown promise in tackling the complexities of 3D head structure in the synthesis of audio-driven talking portraits~\cite{liu2022semantic, guo2021ad}. AD-NeRF~\cite{guo2021ad} firstly presents a NeRF-based talking head framework that directly feeds signals into a conditional implicit function to generate a dynamic neural radiance field. However, static NeRF reconstruction often struggles with scenes featuring non-rigid deformations, resulting in head jitters, motionless mouths, and artifacts.

The inherent limitation of vanilla NeRF in modeling solely static scenes has motivated the development of diverse approaches~\cite{tang2022real, li2023efficient, peng2021neural, peng2021animatable, li2021neural, gafni2021dynamic} to address the representation of dynamic scenes. Notably, deformation-based methods~\cite{tang2022real,li2023efficient} aim to map all observations back to a canonical space by concurrently learning a deformation field alongside the radiance field. RAD-NeRF~\cite{tang2022real} improves inference speed to real-time and introduces grid hash encoding in instant-ngp for talking portrait tasks. ER-NeRF~\cite{li2023efficient} employs Tri-Plane Hash instead of hash encoding in RAD-NeRF, optimizing facial motion effects through region-awareness mechanism. Despite the emergence of many valuable optimizations surrounding NeRF-based talking heads, two persistent issues, visual artifacts and high training costs, continue to hinder the large-scale commercial application of these methods.

\section{Method}

This section details the proposed LokiTalk framework. We begin with the basic preliminaries. Then we elucidate how Region-Specific Deformation Fields establish the fine-grained relationships between driving signals and their corresponding regions. Finally, we explain how ID-Aware Knowledge Transfer enables the extraction of individual characteristics and facilitates the model's learning of shared traits among individuals.

\subsection{Preliminaries and Problem Setting}
Given a set of multi-view images and camera poses, NeRF~\cite{mildenhall2021nerf} represents a static 3D scene with an implicit function $\mathcal{F}:(\mathbf{x},\mathbf{d})\rightarrow (\mathbf{c},\sigma)$, where $\mathbf{x} = (x, y, z)$ is the 3D spatial coordinate and $\mathbf{d} = (\theta, \phi)$ is the viewing direction. The output $\mathbf{c} = (r, g, b)$ is the emitted color and $\sigma$ is the volume density. The color $C(\mathbf{r})$ of one pixel crossed by the ray $\mathbf{r}(t) = \mathbf{o} + t\mathbf{d}$ from camera center $\mathbf{o}$ can be calculated by aggregating the color $\mathbf{c}$ along the ray:
\begin{equation}
\hat{C}(r) = \int^{t_{f}}_{t_{n}}\sigma(\mathbf{r}(t))\cdot \mathbf{c}(\mathbf{r}(t),\mathbf{d})\cdot T(t) dt,
\end{equation}
where $t_{n}$ and $t_{f}$ are the near and far bounds. $T(t)$ is the accumulated transmittance from $t_{n}$ to $t$:
\begin{equation}
T(t) = exp(-\int^{t}_{t_{n}}\sigma(\mathbf(r)(s)) ds).
\end{equation}
Using this fully differentiable volume rendering procedure, NeRF can learn 3D scenes with supervision from only 2D images.

To synthesize in dynamic scenes, an additional condition (\textit{i.e.}, the current time $t$) is required. Previous methods~\cite{park2021nerfies, park2021hypernerf} usually perform dynamic scene modeling via learning a deformation $\Delta \mathbf{x}$ at each position and time step: $\mathcal{G} : \mathbf{x}, t \rightarrow \Delta \mathbf{x}$, which is subsequently added to the original position $\mathbf{x}$. We extend this technique by employing driving signal-conditioned deformation to generate dynamic talking head animations. Other basic settings follow previous NeRF-based works~\cite{guo2021ad, tang2022real, li2023efficient}. 

\subsection{Region-Specific Deformation Fields}

The talking portrait task exhibits two characteristics: (1) Diverse driving signals governing the motion of different regions, and (2) Facial movements indirectly influence torso deformation. Several salient flaws, including mismatched lip synchronization, aberrant blinking, and facial and torso disconnections, can be attributed to the model's inability to establish fine-grained correspondences. Here we describe how to leverage two cascaded deformation fields for hierarchical modeling of driving signals and their corresponding regions.

\paragraph{Driving Signals} 
While speech signals directly drive lip movements, their control over eye blinking is relatively stochastic. This can result in inaccurate lip synchronization and unnatural blinking when directly mapping audio signals to all facial motions~\cite{tang2022real}. To mitigate the interference of audio signals on eye blinking, we computed landmark-based eye aspect ratios  $\mathbf{F}_{\mathbf{e}}$ for explicit control of eye blinking. For other facial regions,  a VAE-based Audio2Motion feature $\mathbf{F}_{\mathbf{a}}$ is used to directly control lip movements and related subtle motions. 

Considering that the torso is primarily driven by the head posture, while simultaneously being influenced by the jaw movements. We use both a 3DMM headpose $\mathbf{F}_{\mathbf{h}}$ and the facial deformation result to predict the torso deformation field. 

To amplify the impact of the driving signal on the affected region, we calculated the cross-attention between the driving signal and the corresponding region, as illustrated in Figure~\ref{fig:framework}.

\paragraph{Deformation Fields}

Given the driving signals $\{ \mathbf{F}_{\mathbf{a}}, \mathbf{F}_{\mathbf{e}},  \mathbf{F}_{\mathbf{h}}\}$, we propose to learn two cascaded deformation fields.

Face deformation module $\Phi_{face}$ is conditioned on $\mathbf{F}_{\mathbf{a}}$ and $\mathbf{F}_{\mathbf{e}}$ to predict the face deformation field:

\begin{equation}
\Delta \mathbf{x}_{face} = \Phi_{face}(PE(\mathbf{x}); \mathbf{F}_{\mathbf{a}}, \mathbf{F}_{\mathbf{e}}),
\end{equation}
where $PE(\cdot)$ is a hash encoder.

Torso deformation filed is predicted with both driving signal $\mathbf{F}_{\mathbf{h}}$ and face deformation $\Delta \mathbf{x}_{face}$ by module $\Phi_{torso}$:

\begin{equation}
\Delta \mathbf{x}_{torso} = \Phi_{torso}(PE(\mathbf{x}); \mathbf{F}_{\mathbf{h}}, \Delta \mathbf{x}_{face}).
\end{equation}

With the learned deformation fields, all observation-space coordinates $\mathbf{x}$ can be warped to the unified canonical-space $\mathbf{x}^{'}$:

\begin{equation}
\mathbf{x}' = \mathbf{x} + \Delta \mathbf{x}_{face} + \Delta \mathbf{x}_{torso}.
\end{equation}

We also introduce a region regularization loss ~\eqref{eq:delta-loss} to optimize cascaded deformation fields, preventing significant coordinate distortions in unrelated regions. This loss function is detailed in the following section.

\begin{figure}[t]
    \centering
    \includegraphics[width=1\linewidth]{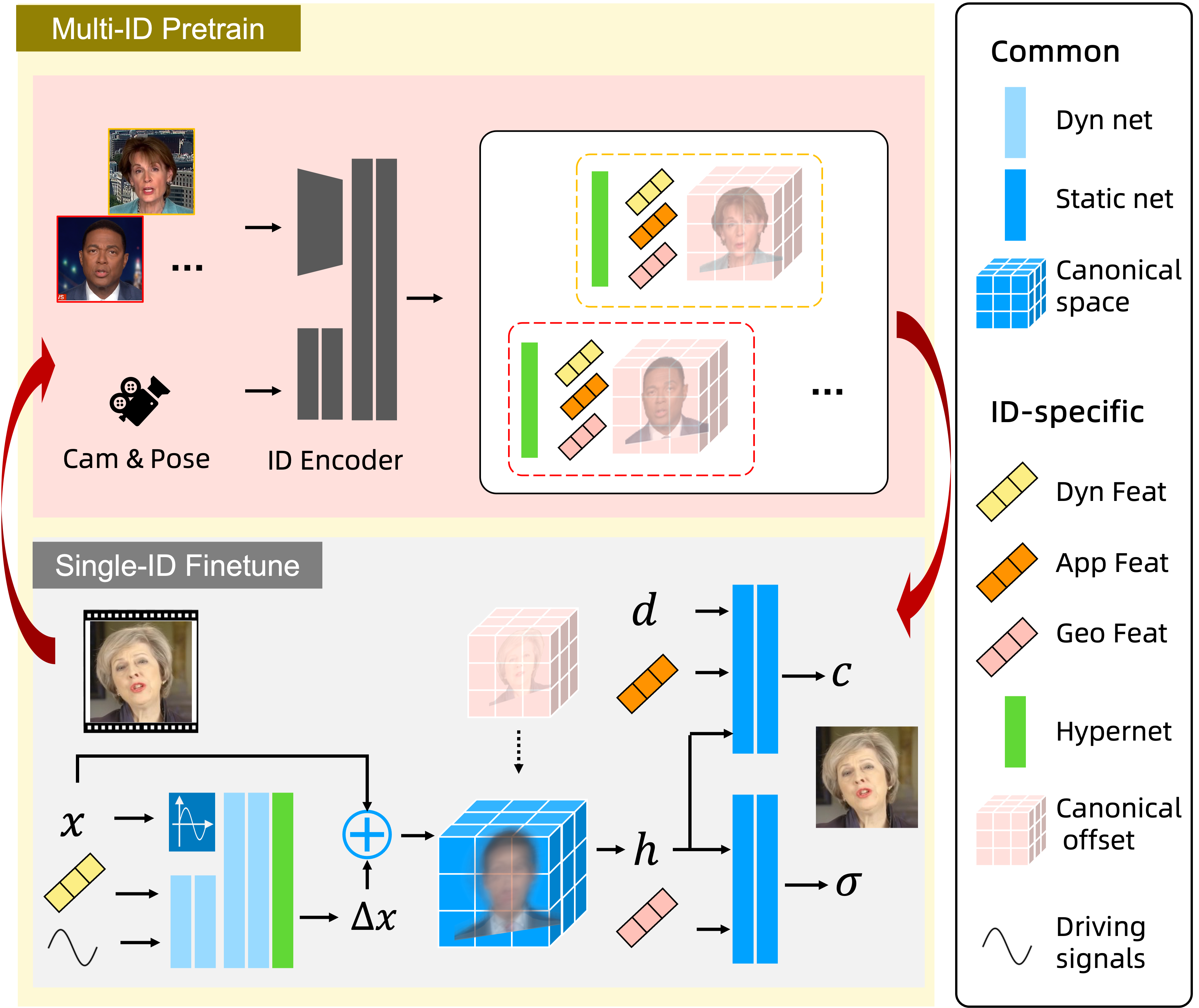}
    \caption{\textbf{ID-Aware Knowledge Transfer.} The blue modules are the common correspondences among multiple identities, comprising dynamic (light blue) and static (dark blue) correspondences. The colored modules are dynamic (facial actions) and static information (geometry and appearance) of individual identities. During the pre-training (entire yellow panel), both upper and lower parts are trained simultaneously on multi-ID data, allowing the model to learn universal information while extracting individual information. When fine-tuning, the lower half will continue training based on the id-aware initialization parameters obtained from the ID-Encoder.
    }
    \label{fig:framework2}
\end{figure}

\begin{figure*}[t]
    \centering
    \includegraphics[width=1\linewidth]{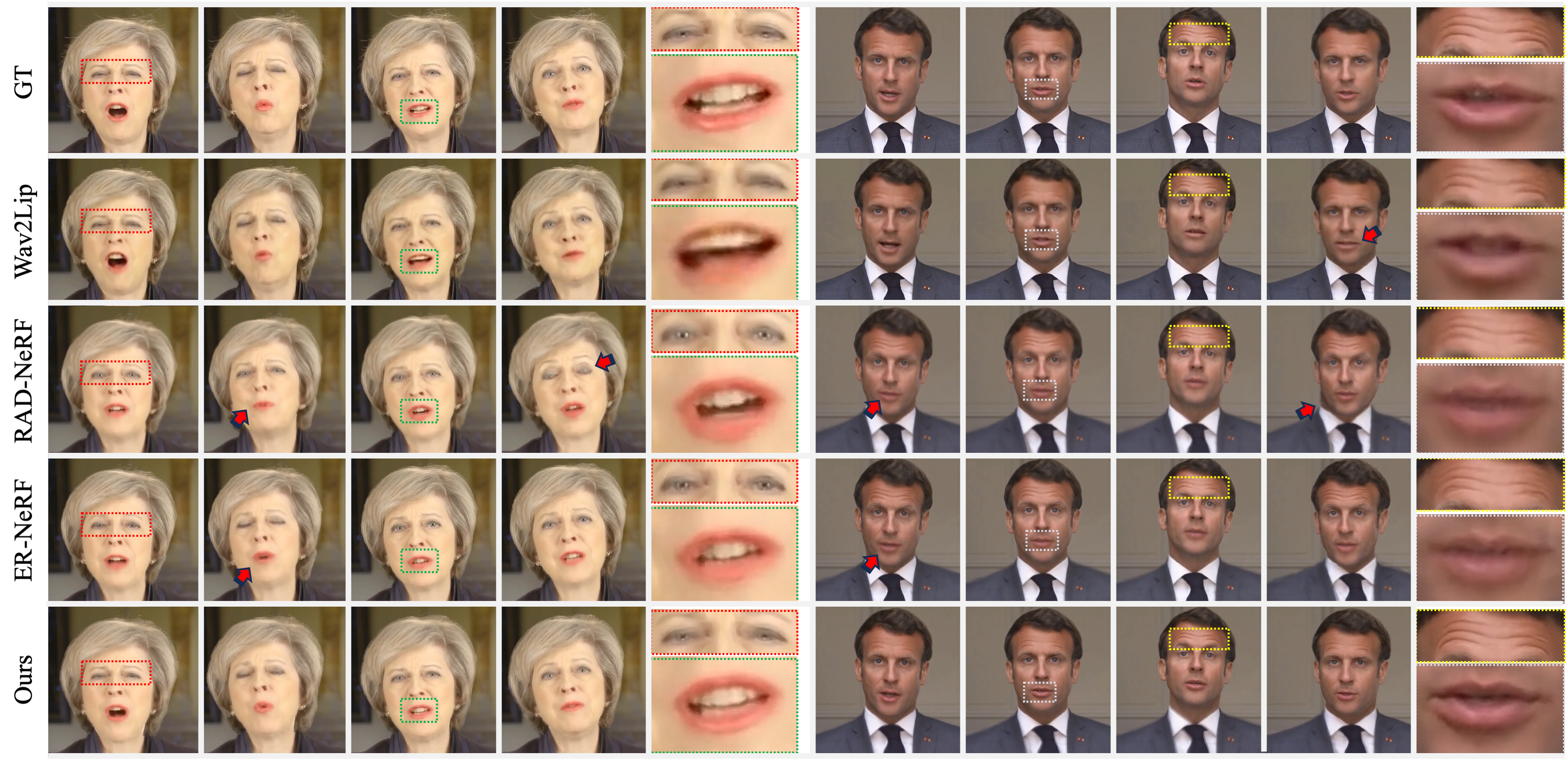}
    \caption{\textbf{The comparison of the keyframes and details of generated portraits.} We mark the un-sync and bad rendering quality results with red arrows, around which the generated eyes, mouths, neck broken or wrinkles are clearly not in line with the real ones. We also show the details of the eyes, teeth, forehead wrinkles and mouth area. Please zoom in for better visualization.}
    \label{fig:res}
\end{figure*}

\subsection{ID-Aware Knowledge Transfer}

Delving into the talking portrait task, different individuals share certain common static (basic head structure) and dynamic representations (basic expressions). Besides, they possess their own ID-specific static (facial and mouth shape, skin color, etc.) and dynamic characteristics (pronunciation articulation, expression intensity, etc.). In light of these observations, we present ID-Aware Knowledge Transfer, a plug-and-play module that learns generalizable static geometric representations and dynamic motion patterns from a limited number of high-quality multi-identity videos, while simultaneously extracting ID-specific static and dynamic features to refine the depiction of individual characters.

\paragraph{Structure} 
As shown in Figure~\ref{fig:framework2}, we jointly trained an ID-Encoder to extract distinctive dynamic information (dynamic features, learnable hyper-networks) and static information (appearance features, geometric features, canonical offset) for each individual. These ID-specific representations are utilized in both the dynamic modeling and static modeling stages of NeRF training.

In the dynamic driving stage, inspired by~\cite{Gafni2020DynamicNR, tang2022real}, we concatenate the identity features with the corresponding driving signal and feed them into MLPs which learn the deformation fields. Inspired by~\cite{zakharov2020fast, wang2019few}, we leverage a learnable hyper-network to enhance the representation of individual dynamic characteristics. We employ the hyper-network to replace the last layer of the MLPs, as this layer has the most direct influence on transferring general knowledge to specific identities.

In the static modeling stage, we incorporate the extracted identity features in both the encoding and decoding modules as the dynamic driving. We overlay id-aware canonical offsets on a shared common canonical space across different individuals, which represents an ID-aware canonical space to encode ID-aware geometric features. We also utilize the ID-aware appearance feature and geometry feature to assist the color decoder and density decoder in reconstructing the appearance and structure specific to each identity.

\begin{table*}[t]
    \resizebox{1\linewidth}{!}{
        \setlength{\tabcolsep}{3mm}
        \centering

        \begin{tabular}{l|cc|ccc|ccc}
        \hline\hline
        
        Method        &  PSNR $\uparrow$  &  LPIPS $\downarrow$  &  LMD $\downarrow$  &  LMD-E $\downarrow$  &  Sync $\uparrow$  &  Cost & FPS & Size(MB) \\
        \hline
        GT  &  $\infty$         &  0                   &  0                 &  0                     &  7.897            &  -      & - & -       \\
        \hline
        Wav2Lip       &  31.148           &  0.074               &  3.069             &  \textbf{2.146}        &  \textbf{8.259}   &  -     & 20 &	\textgreater 400        \\
        AD-NeRF       &  30.413           &  0.081               &  4.315             &  2.407                 &  5.215            &  18h   &  0.11	& \underline{10.54}        \\
        RAD-NeRF      &  31.510           &  0.068               &  3.008             &  2.283                 &  4.410            &  5h  & \underline{32} &	16.47          \\
        GeneFace      & 31.047	          &  0.049               &	\underline{2.903}         	  &  2.248	               &  5.245            & 	10h     & 21	&  48.08   \\
        ER-NeRF       &  \underline{32.506}   &  \underline{0.035}       &  2.917     &  2.219                 &  4.944            &  \textbf{2h}  & \textbf{34}	& \textbf{7.14} \\
        
        \hline
        Ours          &  \textbf{33.744}  &  \textbf{0.029}      &  \textbf{2.732}    &  \underline{2.214}         &  \underline{5.736}    &  \underline{3h}   & 29	& 22.90 \\
        \hline\hline
    \end{tabular}
    
    }
    \setlength{\abovecaptionskip}{1.5mm}
    \caption{\textbf{The quantitative results of the portrait reconstruction}. The \textbf{best} and \underline{second best} results are in \textbf{bold} and \underline{underline} specifically.}
    \label{tab:ev1}
\end{table*}

\paragraph{Training Strategies} 

During pretraining on multi-ID data, the ID-Aware module's input is a random frame from the training identity. The ID-Encoder is trained together with the base model, and the extracted identity features and hyper-networks are synchronously utilized in the base model. By explicitly extracting ID information and applying it to specific network layers, we aim to reduce the network's confusion when being trained on multi-ID data and allow the ID-independent parts of the network to learn general knowledge.

During fine-tuning on a single identity, we only utilize the first frame to pass through the ID-Encoder and obtain the ID-aware canonical offset and hyper-network. The hyper-network becomes regular network weights, while the id-aware canonical offset is added on a shared common canonical space to become a learnable code. Both parts are updated along with other modules during training. After the single frame initialization, the ID-Encoder is discarded. In ablation study, we demonstrate that this plug-and-play module can effectively reduce data dependency and can be applied to other state-of-the-art methods.

\subsection{Loss Function}

We use the weighted MSE loss on each pixel's color $C$ to train our LokiTalk:

\begin{equation}
        L_{color} = \sum_{i \in I}{ w_i \cdot \left\| C_{i} - \hat{C}_{i} \right\| _2^2 },
\end{equation}
where $i$ denotes the $i^{th}$ pixel in image $I$. We observe varying convergence rates across different regions during training and the facial regions with large deformation (\textit{e.g.}, eyes and mouth) are more challenging to converge. This phenomenon can be attributed to stable components being inherently easy to fit. Following this observation, we adapt the learning speed of different regions by constructing a weight matrix $w_{i}$.

For region regularization loss $L_{\Delta}$, we employ the segmentation algorithm to partition the portrait into face and torso, providing corresponding mask matrices $w_{\mathbf{x}_{face}}$ and $w_{\mathbf{x}_{torso}}$. In matrix $w_{\mathbf{x}_{face}}$, the face area is assigned the value of 1, with all other positions set to 0. Conversely, matrix $w_{\mathbf{x}_{torso}}$ assigns the weight of 1 to the torso, while the remaining positions are set to 0:

\begin{equation}
    \label{eq:delta-loss}
    \begin{split}
        L_{\Delta} =  \sum  ( \left \| \Delta \mathbf{x}_{face} \right \| \cdot (1 - w_{\mathbf{x}_{face}}) \\
        + \left \| \Delta \mathbf{x}_{torso} \right \| \cdot (1 - w_{\mathbf{x}_{torso}})  ).
    \end{split}
\end{equation}

In order to enhance the effect of signal-region cross-attention, we use attention regularization loss to punish signal-irrelevant areas:
\begin{equation}
    L_{att-*} = \sum \| f_{*} \cdot (1 - m_{*})\|,
\end{equation}
where $f_{*}$ is the attention score and $m_{*}$ is the corresponding region mask. The final attention regularization loss is:
\begin{equation}
    L_{att} = L_{att-eye} + L_{att-lip} + L_{att-torso}
\end{equation}

Besides, an entropy regularization term~\cite{tang2022real} is used to encourage the pixel transparency to be either 0 or 1:
\begin{equation}
    \begin{aligned}
        L_{\alpha} = - \sum_{\alpha \in I}{ \left ( \alpha \log{\alpha} + (1 - \alpha) \log{(1 - \alpha)} \right ) }.
    \end{aligned}
\end{equation}

Similar to~\cite{li2023efficient}, we also introduce the LPIPS loss to enhance the realness:
\begin{equation}
    \begin{aligned}
        L_{lpips} = \text{LPIPS} (I, \hat{I}).
    \end{aligned}
\end{equation}

Overall, we supervise the training in pixel space, 3D space, and feature space jointly, as well as apply regularization to assist in optimization:
\begin{equation}
    \begin{split}
        L = L_{color} + \lambda_{\Delta} \cdot L_{\Delta} + \lambda_{att} \cdot L_{att} \\
        + \lambda_{\alpha} \cdot L_{\alpha} + \lambda_{lpips} \cdot L_{lpips},
    \end{split}
\end{equation}
where $\lambda_{\Delta}$, $\lambda_{att}$, $\lambda_{\alpha}$ and $ \lambda_{lpips}$ are weight coefficients.

\begin{figure}[t]
    \centering
    \includegraphics[width=1\linewidth]{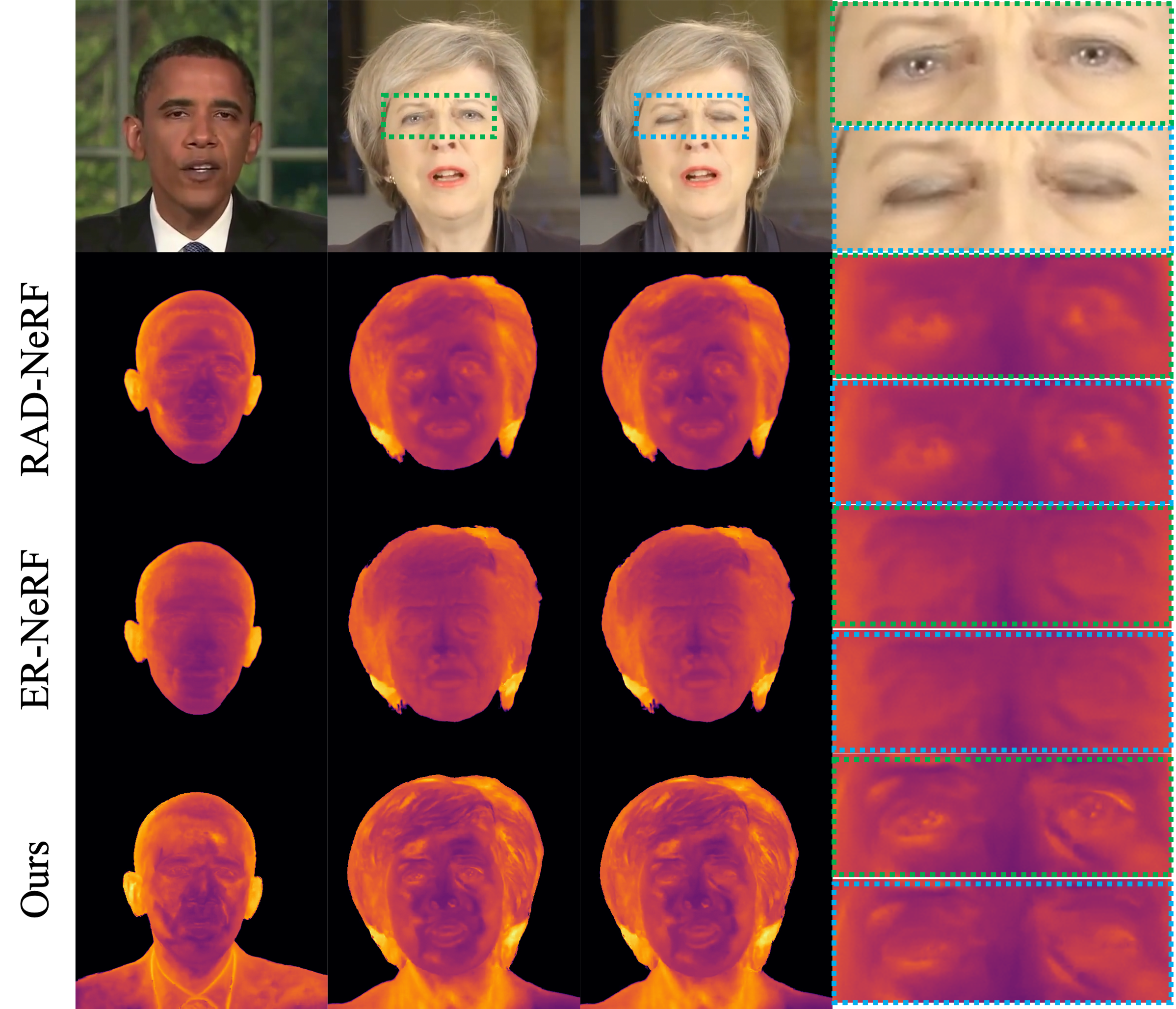}
    \caption{\textbf{Comparison of the depth maps generated by our method and the baseline methods.} Our depth map shows more details on the face area, especially the mouth and eye expressions (differences between open and closed eyes). The connection between the face and torso is more consistent in our results.}
    \label{fig:depth}
\end{figure}

\section{Experiment}

\paragraph{Dataset and Data Preprocessing.} 
We use the datasets collected by previous works~\cite{guo2021ad, lu2021lsp, shen2022learning, li2023efficient} for our experiments. Each video has an average duration of 4 minutes. Similar to AD-NeRF~\cite{guo2021ad}, we process facial parsing and 3DMM face tracking on each video to obtain training images, camera intrinsic, and head poses. Each video is divided into training and validation sets at a ratio of 10:1.

\paragraph{Comparison Settings and Metrics.}
We compare our method with a one-shot approach, Wav2Lip~\cite{prajwal2020lip}, and three end-to-end NeRF-based models: AD-NeRF~\cite{guo2021ad}, RAD-NeRF~\cite{tang2022real}, and ER-NeRF~\cite{li2023efficient}. To facilitate the transparent comparison, we evaluate our method directly on the Ground Truth. All these approaches are implemented using their official code. For metrics, we employ Peak Signal-to-Noise Ratio (\textbf{PSNR}) to measure the overall image quality, and Learned Perceptual Image Patch Similarity (\textbf{LPIPS})~\cite{zhang2018lpips} to measure the details. We utilize the landmark distance (\textbf{LMD} for face, \textbf{LMD-E} for eye)~\cite{Chen2018LipMG} and SyncNet confidence score (\textbf{Sync})~\cite{chung2017syncnet2} to measure the face motion accuracy.

\begin{figure}[t]
    \centering
    \includegraphics[width=1\linewidth]{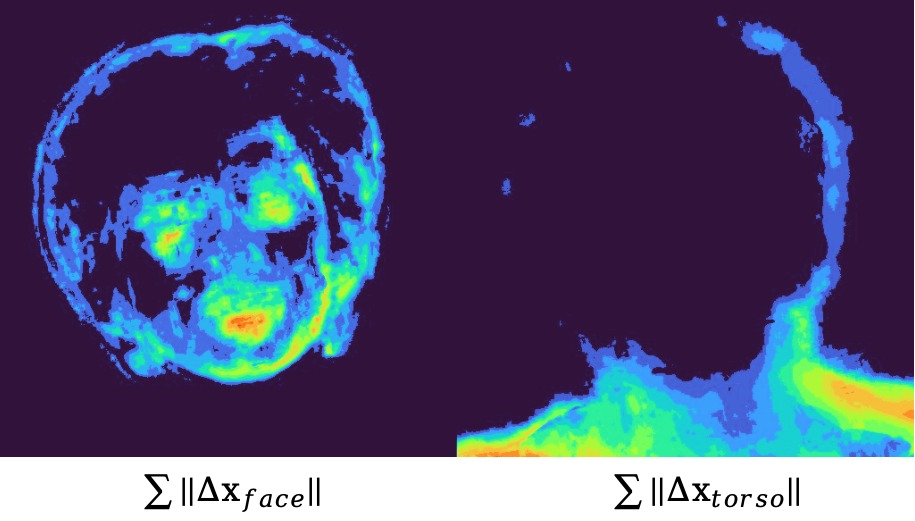}

    \caption{\textbf{Heatmaps of the $\sum{ \left \| \Delta \mathbf{x}_{face} \right \|}$ and $\sum{ \left \| \Delta \mathbf{x}_{torso} \right \|}$.} Brighter areas represent regions with more dynamic deformations. The reason for the bright area close to the hair edges is due to the jitter in parsing results which mislead learning of the deformations.}
    \label{fig:heatmap}
\end{figure}

\begin{table}[t]
    \resizebox{1\linewidth}{!}{
        \setlength{\tabcolsep}{1.5mm}
        \centering
        \begin{tabular}{l|cccc}
        \hline\hline
        
        Method        &  PSNR $\uparrow$  &  LPIPS $\downarrow$  &  LMD $\downarrow$  &  Sync $\uparrow$   \\
        \hline
        O+D+R&  30.464&  0.051&  3.132& 5.423\\
        O+D&  29.203&  0.055&  3.299& 4.728\\
        O&  28.125&  0.074&  3.592& 3.410\\
        \hline\hline
        \end{tabular}
    }
    \setlength{\abovecaptionskip}{1.5mm}
    \caption{\textbf{Ablation Study on different modules.} \textbf{O} is the original structure, like RAD-NeRF, \textbf{O+D} is the original model with the dynamic module, and \textbf{O+D+R} is our base model (with Region-Specific Deformation Field Learning Module).}
    \label{tab:t2}
\end{table}

\paragraph{Implementation Details.} We implement our framework in PyTorch. 
In the pretraining stage, the ID-Encoder and base model are jointly trained. For each step, we randomly choose a reference frame in the same video as input. The model is trained for 50 epochs with the initial learning rate of $1\times 10^{-3}$ and decreases exponentially to $1\times 10^{-4}$ at last. 
For a specific identity, we first initialize the model from the first training frame by the pre-trained ID-Encoder and then the ID-Encoder can be discarded. Then we treat the ID embeddings as learnable parameters that update with other modules. In this stage, we train the model for 10 epochs with the initial learning rate of $1\times 10^{-3}$ for ID embeddings and $1\times 10^{-4}$ for other modules and decreases exponentially to $1\times 10^{-4}$ and $1\times 10^{-5}$ respectively. 
In both stage, for the first 80\% iterations, $\lambda_{lpips}=0$, and for the remaining 20\% iterations, $\lambda_{lpips}=5 \times 10^{-3}$. In all iterations $\lambda_{\alpha} = 1 \times 10^{-4}$, $\lambda_{\Delta}=1 \times 10^{-5}$.
All experiments are conducted on a single NVIDIA Tesla A100 GPU.

\subsection{Quantitative Results}

\paragraph{Evaluation Results.}
The quantitative evaluation results are shown in Table~\ref{tab:ev1}.
Our method performs the best in most metrics, it is evident that LokiTalk demonstrates the best overall performance among the presented models,  underscoring its capability for faithful reconstruction. Our method performs clearly better than the state-of-the-art model ER-NeRF, showing that the deformation field is useful for modeling dynamic scenes and jointly learning the whole portrait is beneficial for obtaining realistic results. 
It is worth mentioning that Wav2Lip achieves the best LMD-E and Sync metrics, with the Sync value even surpassing the ground truth. This is because the Syncnet feature is incorporated into the training of Wav2Lip, and the real eye region serves as its input during inference.
We also evaluate the results of cross-driving, which are presented in the supplementary material.

\begin{table*}[t]
    \resizebox{1\linewidth}{!}{
        \setlength{\tabcolsep}{3mm}
        \centering
        \begin{tabular}{l|cccccc|c}
        \hline\hline
        
        Pretrain IDs        &  0   & 10        & 10        & 10        & 5         & 3              &\multirow{2}{*}{ER-NeRF}\\
        Finetune Data       &  100\%  & 100\%  & 50\%      & 25\%      & 50\%      & 50\%             \\
        \hline
        LMD $\downarrow$    & 3.132     & 3.034     & 3.125     & 3.209     & 3.134     & 3.148  &3.334   \\
        Sync $\uparrow$     & 5.423     & 5.873     & 5.569     & 5.396     & 5.404     & 5.386  &4.409   \\
        PSNR $\uparrow$     & 30.464    & 30.482    & 30.387    & 29.832    & 30.387    & 30.153 &28.425   \\
        LPIPS $\downarrow$  & 0.051     & 0.050     & 0.058     & 0.061     & 0.060     & 0.059  &0.064   \\

        \hline\hline
        \end{tabular}
    }
    \setlength{\abovecaptionskip}{1.5mm}
    \caption{\textbf{Ablation Study on ID-Aware Knowledge Transfer.} By leveraging pre-training with 10 IDs, LokiTalk achieves performance surpassing ER-NeRF by using only 25\% of the data. The 100\% finetune data (single id) is a 5-minutes video.}
    \label{tab:t3-1}
\end{table*}

\subsection{Qualitative Results}
\label{sec:qr2}
\paragraph{Evaluation Results.}

For an intuitive comparison of the portrait synthesis effect, we present keyframes of the synthesized video details in Figure~\ref{fig:res}. The zoomed-in details showcase the eyes, teeth, forehead, and mouth.
Although Wav2Lip exhibits decent lip-sync accuracy, the synthesized images are not clear enough and lack realism. RAD-NeRF and ER-NeRF 
are unable to synthesize accurate facial expressions occasionally, resulting in low lip-sync accuracy and incorrect eye blinking, especially in cases where the eyes are half-closed. Additionally, RAD-NeRF demonstrates inconsistencies in head and body movements (broken neck).
Furthermore, in Figure~\ref{fig:depth}, we display the depth maps of the NeRF-based methods. Our method captures more details in depth. In contrast, the depth of RAD-NeRF and ER-NeRF appear smoother in the facial region (first column). When zoomed in, it is evident that our method provides better differentiation in the depth for open and closed eyes, indicating more accurate modeling. Our depth map also shows that the connection between the face and the torso is more consistent. 
Overall, our method delivers high image quality, rich facial details (clearer teeth and more realistic forehead wrinkles), accurate expression rendering (lip-sync, eye blinking), and the ability to synthesize realistic portrait results.

\subsection{Ablation Study}
\label{sec:as}
\paragraph{Region-Specific Deformation fields}

In Table~\ref{tab:t2}, by removing the region-specific designing, a notable decline is observed, showing that the utilization of distinct signals to guide motions in specific areas can significantly improve the performance of the model. We also visualize the heatmaps of the face and torso displacement field in Figure~\ref{fig:heatmap}, and find that jointly optimized methods can effectively learn overall expressions, mitigating issues such as torso fractures. Further, we analyze the unified canonical space to reveal that there is a serious performance degradation in the reconstruction if we remove this stage. This result proves that deformation fields are beneficial for modeling dynamic scenes.

\paragraph{ID-Aware Knowledge Transfer.}

In Table~\ref{tab:t3-1} and Table~\ref{tab:t4-1}, we experiment with different settings of the ID-Aware knowledge transfer method to showcase its effectiveness and generality. With an increase in the number of auxiliary training videos, the performance continues to improve. It is noteworthy that when reducing the duration of specific person videos, this module can also achieve compelling results. To assess the generality of ID-Aware knowledge transfer, we integrate it into the ER-NeRF and evaluate its performance. Specifically, by utilizing only a quarter of the specific video and incorporating 10 auxiliary videos, LokiTalk surpasses the performance of ER-NeRF. As shown in Table~\ref{tab:t4-1}, while the transfer method does not help the ER-NeRF method much under the existing data settings, it demonstrates significant performance enhancements in more challenging data scenarios, such as shorter target videos.

\begin{table}[t]
    \resizebox{1\linewidth}{!}{
        \setlength{\tabcolsep}{1.5mm}
        \centering
        \begin{tabular}{l|cccc}
        \hline\hline
        
        Pretrain IDs        &  0   & 0        & 10        & 10                \\
        Finetune Data       &  100\%  & 50\%  & 100\%      & 50\%                \\
        \hline
        LMD $\downarrow$    & 3.334     & 3.424     & 3.308     & 3.329         \\
        Sync $\uparrow$     & 4.409     & 4.364     & 4.617     & 4.405         \\
        PSNR $\uparrow$     & 28.425    & 27.761    & 28.428    & 28.426       \\
        LPIPS $\downarrow$  & 0.064     & 0.067     & 0.064     & 0.065          \\

        \hline\hline
        \end{tabular}
    }
    \setlength{\abovecaptionskip}{1.5mm}
    \caption{\textbf{The effect of applying our ID-Aware Knowledge Transfer on ER-NeRF.} The results indicate that it is plug-and-play and can enhance the performance of existing methods like ER-NeRF.}
    \label{tab:t4-1}
\end{table}

\section{Conclusion}

In this paper, we propose LokiTalk for dynamic talking portrait synthesis. The Region-Specific Deformation Fields establish a fine-grained correspondence between the driving signals and the affected regions, resulting in more realistic synthesis results. The ID-Aware Knowledge Transfer enables the learning of shared knowledge from multi-ID data and transferring it to specific ID during fine-tuning, reducing data requirements and accelerating training. The proposed method has been successfully applied in enterprise-level scenarios, supporting the large-scale production of high-quality digital avatars.

\normalsize
\bibliography{main}

\begin{thebibliography}{35}
\providecommand{\natexlab}[1]{#1}
\providecommand{\url}[1]{\texttt{#1}}
\expandafter\ifx\csname urlstyle\endcsname\relax
  \providecommand{\doi}[1]{doi: #1}\else
  \providecommand{\doi}{doi: \begingroup \urlstyle{rm}\Url}\fi

\bibitem[Kumar et~al.(2017)Kumar, Sotelo, Kumar, De~Brebisson, and Bengio]{kumar2017obamanet}
Rithesh Kumar, Jose Sotelo, Kundan Kumar, Alexandre De~Brebisson, and Yoshua Bengio.
\newblock Obamanet: Photo-realistic lip-sync from text.
\newblock \emph{arXiv preprint arXiv:1801.01442}, 2017.

\bibitem[Tian et~al.(2019)Tian, Yuan, and Liu]{tian2019audio2face}
Guanzhong Tian, Yi~Yuan, and Yong Liu.
\newblock Audio2face: Generating speech/face animation from single audio with attention-based bidirectional lstm networks.
\newblock In \emph{2019 IEEE International Conference on Multimedia \& Expo Workshops (ICMEW)}, pages 366--371. IEEE, 2019.

\bibitem[Zheng et~al.(2021{\natexlab{a}})Zheng, Zhu, Song, and Ji]{zheng2021neural}
Ruobing Zheng, Zhou Zhu, Bo~Song, and Changjiang Ji.
\newblock A neural lip-sync framework for synthesizing photorealistic virtual news anchors.
\newblock In \emph{2020 25th International Conference on Pattern Recognition (ICPR)}, pages 5286--5293. IEEE, 2021{\natexlab{a}}.

\bibitem[Hong et~al.(2022)Hong, Zhang, Shen, and Xu]{hong2022depth}
Fa-Ting Hong, Longhao Zhang, Li~Shen, and Dan Xu.
\newblock Depth-aware generative adversarial network for talking head video generation.
\newblock In \emph{Proceedings of the IEEE/CVF conference on computer vision and pattern recognition}, pages 3397--3406, 2022.

\bibitem[Zhang et~al.(2023)Zhang, Cun, Wang, Zhang, Shen, Guo, Shan, and Wang]{zhang2023sadtalker}
Wenxuan Zhang, Xiaodong Cun, Xuan Wang, Yong Zhang, Xi~Shen, Yu~Guo, Ying Shan, and Fei Wang.
\newblock Sadtalker: Learning realistic 3d motion coefficients for stylized audio-driven single image talking face animation.
\newblock In \emph{Proceedings of the IEEE/CVF Conference on Computer Vision and Pattern Recognition}, pages 8652--8661, 2023.

\bibitem[Cheng et~al.(2022)Cheng, Cun, Zhang, Xia, Yin, Zhu, Wang, Wang, and Wang]{cheng2022videoretalking}
Kun Cheng, Xiaodong Cun, Yong Zhang, Menghan Xia, Fei Yin, Mingrui Zhu, Xuan Wang, Jue Wang, and Nannan Wang.
\newblock Videoretalking: Audio-based lip synchronization for talking head video editing in the wild.
\newblock In \emph{SIGGRAPH Asia 2022 Conference Papers}, pages 1--9, 2022.

\bibitem[Tian et~al.(2024)Tian, Wang, Zhang, and Bo]{tian2024emo}
Linrui Tian, Qi~Wang, Bang Zhang, and Liefeng Bo.
\newblock Emo: Emote portrait alive-generating expressive portrait videos with audio2video diffusion model under weak conditions.
\newblock \emph{arXiv preprint arXiv:2402.17485}, 2024.

\bibitem[Chen et~al.(2020)Chen, Cui, Liu, Li, Kou, Xu, and Xu]{chen2020talking}
Lele Chen, Guofeng Cui, Celong Liu, Zhong Li, Ziyi Kou, Yi~Xu, and Chenliang Xu.
\newblock Talking-head generation with rhythmic head motion.
\newblock In \emph{European Conference on Computer Vision}, pages 35--51. Springer, 2020.

\bibitem[Suwajanakorn et~al.(2017)Suwajanakorn, Seitz, and Kemelmacher-Shlizerman]{suwajanakorn2017synthesizing}
Supasorn Suwajanakorn, Steven~M Seitz, and Ira Kemelmacher-Shlizerman.
\newblock Synthesizing obama: learning lip sync from audio.
\newblock \emph{ACM Transactions on Graphics (ToG)}, 36\penalty0 (4):\penalty0 1--13, 2017.

\bibitem[Thies et~al.(2020)Thies, Elgharib, Tewari, Theobalt, and Nie{\ss}ner]{thies2020neural}
Justus Thies, Mohamed Elgharib, Ayush Tewari, Christian Theobalt, and Matthias Nie{\ss}ner.
\newblock Neural voice puppetry: Audio-driven facial reenactment.
\newblock In \emph{Computer Vision--ECCV 2020: 16th European Conference, Glasgow, UK, August 23--28, 2020, Proceedings, Part XVI 16}, pages 716--731. Springer, 2020.

\bibitem[Zheng et~al.(2021{\natexlab{b}})Zheng, Song, and Ji]{zheng2021learning}
Ruobing Zheng, Bo~Song, and Changjiang Ji.
\newblock Learning pose-adaptive lip sync with cascaded temporal convolutional network.
\newblock In \emph{ICASSP 2021-2021 IEEE International Conference on Acoustics, Speech and Signal Processing (ICASSP)}, pages 4255--4259. IEEE, 2021{\natexlab{b}}.

\bibitem[Prajwal et~al.(2020)Prajwal, Mukhopadhyay, Namboodiri, and Jawahar]{prajwal2020lip}
KR~Prajwal, Rudrabha Mukhopadhyay, Vinay~P Namboodiri, and CV~Jawahar.
\newblock A lip sync expert is all you need for speech to lip generation in the wild.
\newblock In \emph{Proceedings of the 28th ACM international conference on multimedia}, pages 484--492, 2020.

\bibitem[Zhou et~al.(2019)Zhou, Liu, Liu, Luo, and Wang]{zhou2019talking}
Hang Zhou, Yu~Liu, Ziwei Liu, Ping Luo, and Xiaogang Wang.
\newblock Talking face generation by adversarially disentangled audio-visual representation.
\newblock In \emph{Proceedings of the AAAI Conference on Artificial Intelligence}, volume~33, pages 9299--9306, 2019.

\bibitem[Peng et~al.(2024)Peng, Hu, Shi, Zhu, Zhang, Zhao, He, Liu, and Fan]{peng2024synctalk}
Ziqiao Peng, Wentao Hu, Yue Shi, Xiangyu Zhu, Xiaomei Zhang, Hao Zhao, Jun He, Hongyan Liu, and Zhaoxin Fan.
\newblock Synctalk: The devil is in the synchronization for talking head synthesis.
\newblock In \emph{Proceedings of the IEEE/CVF Conference on Computer Vision and Pattern Recognition}, pages 666--676, 2024.

\bibitem[Mildenhall et~al.(2021)Mildenhall, Srinivasan, Tancik, Barron, Ramamoorthi, and Ng]{mildenhall2021nerf}
Ben Mildenhall, Pratul~P Srinivasan, Matthew Tancik, Jonathan~T Barron, Ravi Ramamoorthi, and Ren Ng.
\newblock Nerf: Representing scenes as neural radiance fields for view synthesis.
\newblock \emph{Communications of the ACM}, 65\penalty0 (1):\penalty0 99--106, 2021.

\bibitem[Guo et~al.(2021)Guo, Chen, Liang, Liu, Bao, and Zhang]{guo2021ad}
Yudong Guo, Keyu Chen, Sen Liang, Yong-Jin Liu, Hujun Bao, and Juyong Zhang.
\newblock Ad-nerf: Audio driven neural radiance fields for talking head synthesis.
\newblock In \emph{Proceedings of the IEEE/CVF International Conference on Computer Vision}, pages 5784--5794, 2021.

\bibitem[Tang et~al.(2022)Tang, Wang, Zhou, Chen, He, Hu, Liu, Zeng, and Wang]{tang2022real}
Jiaxiang Tang, Kaisiyuan Wang, Hang Zhou, Xiaokang Chen, Dongliang He, Tianshu Hu, Jingtuo Liu, Gang Zeng, and Jingdong Wang.
\newblock Real-time neural radiance talking portrait synthesis via audio-spatial decomposition.
\newblock \emph{arXiv preprint arXiv:2211.12368}, 2022.

\bibitem[Li et~al.(2023)Li, Zhang, Bai, Zhou, and Gu]{li2023efficient}
Jiahe Li, Jiawei Zhang, Xiao Bai, Jun Zhou, and Lin Gu.
\newblock Efficient region-aware neural radiance fields for high-fidelity talking portrait synthesis.
\newblock In \emph{Proceedings of the IEEE/CVF International Conference on Computer Vision}, pages 7568--7578, 2023.

\bibitem[Yao et~al.(2022)Yao, Zhong, Yan, Zhai, and Yang]{yao2022dfa}
Shunyu Yao, RuiZhe Zhong, Yichao Yan, Guangtao Zhai, and Xiaokang Yang.
\newblock Dfa-nerf: Personalized talking head generation via disentangled face attributes neural rendering.
\newblock \emph{arXiv preprint arXiv:2201.00791}, 2022.

\bibitem[Liu et~al.(2022)Liu, Xu, Wu, Zhou, Wu, and Zhou]{liu2022semantic}
Xian Liu, Yinghao Xu, Qianyi Wu, Hang Zhou, Wayne Wu, and Bolei Zhou.
\newblock Semantic-aware implicit neural audio-driven video portrait generation.
\newblock In \emph{European Conference on Computer Vision}, pages 106--125. Springer, 2022.

\bibitem[Shen et~al.(2022)Shen, Li, Zhu, Duan, Zhou, and Lu]{shen2022learning}
Shuai Shen, Wanhua Li, Zheng Zhu, Yueqi Duan, Jie Zhou, and Jiwen Lu.
\newblock Learning dynamic facial radiance fields for few-shot talking head synthesis.
\newblock In \emph{European Conference on Computer Vision}, pages 666--682. Springer, 2022.

\bibitem[Zhang et~al.(2024)Zhang, Zheng, Li, Han, Li, Wang, Guo, Chen, Liu, and Yang]{zhang2024learning}
Zicheng Zhang, Ruobing Zheng, Bonan Li, Congying Han, Tianqi Li, Meng Wang, Tiande Guo, Jingdong Chen, Ziwen Liu, and Ming Yang.
\newblock Learning dynamic tetrahedra for high-quality talking head synthesis.
\newblock In \emph{Proceedings of the IEEE/CVF Conference on Computer Vision and Pattern Recognition}, pages 5209--5219, 2024.

\bibitem[Peng et~al.(2021{\natexlab{a}})Peng, Zhang, Xu, Wang, Shuai, Bao, and Zhou]{peng2021neural}
Sida Peng, Yuanqing Zhang, Yinghao Xu, Qianqian Wang, Qing Shuai, Hujun Bao, and Xiaowei Zhou.
\newblock Neural body: Implicit neural representations with structured latent codes for novel view synthesis of dynamic humans.
\newblock In \emph{Proceedings of the IEEE/CVF Conference on Computer Vision and Pattern Recognition}, pages 9054--9063, 2021{\natexlab{a}}.

\bibitem[Peng et~al.(2021{\natexlab{b}})Peng, Dong, Wang, Zhang, Shuai, Zhou, and Bao]{peng2021animatable}
Sida Peng, Junting Dong, Qianqian Wang, Shangzhan Zhang, Qing Shuai, Xiaowei Zhou, and Hujun Bao.
\newblock Animatable neural radiance fields for modeling dynamic human bodies.
\newblock In \emph{Proceedings of the IEEE/CVF International Conference on Computer Vision}, pages 14314--14323, 2021{\natexlab{b}}.

\bibitem[Li et~al.(2021)Li, Niklaus, Snavely, and Wang]{li2021neural}
Zhengqi Li, Simon Niklaus, Noah Snavely, and Oliver Wang.
\newblock Neural scene flow fields for space-time view synthesis of dynamic scenes.
\newblock In \emph{Proceedings of the IEEE/CVF Conference on Computer Vision and Pattern Recognition}, pages 6498--6508, 2021.

\bibitem[Gafni et~al.(2021)Gafni, Thies, Zollhofer, and Nie{\ss}ner]{gafni2021dynamic}
Guy Gafni, Justus Thies, Michael Zollhofer, and Matthias Nie{\ss}ner.
\newblock Dynamic neural radiance fields for monocular 4d facial avatar reconstruction.
\newblock In \emph{Proceedings of the IEEE/CVF Conference on Computer Vision and Pattern Recognition}, pages 8649--8658, 2021.

\bibitem[Park et~al.(2021{\natexlab{a}})Park, Sinha, Barron, Bouaziz, Goldman, Seitz, and Martin-Brualla]{park2021nerfies}
Keunhong Park, Utkarsh Sinha, Jonathan~T. Barron, Sofien Bouaziz, Dan~B Goldman, Steven~M. Seitz, and Ricardo Martin-Brualla.
\newblock Nerfies: Deformable neural radiance fields.
\newblock \emph{ICCV}, 2021{\natexlab{a}}.

\bibitem[Park et~al.(2021{\natexlab{b}})Park, Sinha, Hedman, Barron, Bouaziz, Goldman, Martin-Brualla, and Seitz]{park2021hypernerf}
Keunhong Park, Utkarsh Sinha, Peter Hedman, Jonathan~T Barron, Sofien Bouaziz, Dan~B Goldman, Ricardo Martin-Brualla, and Steven~M Seitz.
\newblock Hypernerf: A higher-dimensional representation for topologically varying neural radiance fields.
\newblock \emph{arXiv preprint arXiv:2106.13228}, 2021{\natexlab{b}}.

\bibitem[Gafni et~al.(2020)Gafni, Thies, Zollhofer, and Nie{\ss}ner]{Gafni2020DynamicNR}
Guy Gafni, Justus Thies, Michael Zollhofer, and Matthias Nie{\ss}ner.
\newblock Dynamic neural radiance fields for monocular 4d facial avatar reconstruction.
\newblock \emph{2021 IEEE/CVF Conference on Computer Vision and Pattern Recognition (CVPR)}, pages 8645--8654, 2020.
\newblock URL \url{https://api.semanticscholar.org/CorpusID:227342468}.

\bibitem[Zakharov et~al.(2020)Zakharov, Ivakhnenko, Shysheya, and Lempitsky]{zakharov2020fast}
Egor Zakharov, Aleksei Ivakhnenko, Aliaksandra Shysheya, and Victor Lempitsky.
\newblock Fast bi-layer neural synthesis of one-shot realistic head avatars.
\newblock In \emph{Computer Vision--ECCV 2020: 16th European Conference, Glasgow, UK, August 23--28, 2020, Proceedings, Part XII 16}, pages 524--540. Springer, 2020.

\bibitem[Wang et~al.(2019)Wang, Liu, Tao, Liu, Kautz, and Catanzaro]{wang2019few}
Ting-Chun Wang, Ming-Yu Liu, Andrew Tao, Guilin Liu, Jan Kautz, and Bryan Catanzaro.
\newblock Few-shot video-to-video synthesis.
\newblock \emph{arXiv preprint arXiv:1910.12713}, 2019.

\bibitem[Lu et~al.(2021)Lu, Chai, and Cao]{lu2021lsp}
Yuanxun Lu, Jinxiang Chai, and Xun Cao.
\newblock Live speech portraits: Real-time photorealistic talking-head animation.
\newblock \emph{ACM Trans. Graph.}, 40\penalty0 (6), dec 2021.
\newblock ISSN 0730-0301.
\newblock \doi{10.1145/3478513.3480484}.
\newblock URL \url{https://doi.org/10.1145/3478513.3480484}.

\bibitem[Zhang et~al.(2018)Zhang, Isola, Efros, Shechtman, and Wang]{zhang2018lpips}
Richard Zhang, Phillip Isola, Alexei~A Efros, Eli Shechtman, and Oliver Wang.
\newblock The unreasonable effectiveness of deep features as a perceptual metric.
\newblock In \emph{Proceedings of the IEEE conference on computer vision and pattern recognition}, pages 586--595, 2018.

\bibitem[Chen et~al.(2018)Chen, Li, Maddox, Duan, and Xu]{Chen2018LipMG}
Lele Chen, Zhiheng Li, Ross~K. Maddox, Zhiyao Duan, and Chenliang Xu.
\newblock Lip movements generation at a glance.
\newblock \emph{ArXiv}, abs/1803.10404, 2018.
\newblock URL \url{https://api.semanticscholar.org/CorpusID:4435268}.

\bibitem[Chung and Zisserman(2017)]{chung2017syncnet2}
Joon~Son Chung and Andrew Zisserman.
\newblock Out of time: Automated lip sync in the wild.
\newblock In \emph{Computer Vision--ACCV 2016 Workshops: ACCV 2016 International Workshops, Taipei, Taiwan, November 20-24, 2016, Revised Selected Papers, Part II 13}, pages 251--263. Springer, 2017.

\end{thebibliography}


\end{document}